\pdfoutput=1
\documentclass[letterpaper,journal]{IEEEtran}
\usepackage{amsmath,amsfonts,amssymb}
\usepackage{algorithmic}
\usepackage{algorithm}
\usepackage[linesnumbered,ruled,vlined,algo2e]{algorithm2e}
\usepackage{array}
\usepackage[caption=false,font=normalsize,labelfont=sf,textfont=sf]{subfig}
\usepackage{textcomp}
\usepackage{stfloats}
\usepackage{url}
\usepackage{verbatim}
\usepackage{graphicx}
\usepackage[numbers,sort&compress]{natbib}
\usepackage{booktabs}
\usepackage{multirow}
\usepackage{tabularx}
\usepackage{xcolor}
\graphicspath{{figures/}}
\newcommand{\vu}{\mathbf{u}}
\newcommand{\vx}{\mathbf{x}}
\newcommand{\vz}{\mathbf{z}}
\newcommand{\vo}{\mathbf{o}}
\newcommand{\vc}{\mathbf{c}}
\newcommand{\vg}{\mathbf{g}}
\newcommand{\vh}{\mathbf{h}}
\newcommand{\vs}{\mathbf{s}}
\newcommand{\vM}{\mathbf{M}}
\newcommand{\vp}{\mathbf{p}}
\newcommand{\vth}{\boldsymbol{\theta}}
\newcommand{\dxi}{\Delta\xi}
\newcommand{\Geo}{\mathcal{G}}
\newcommand{\Uth}{\mathcal{U}_{\vth}}
\newcommand{\method}{Hourglass}
\newtheorem{remark}{Remark}

\usepackage[colorlinks=true,linkcolor=blue,citecolor=blue,urlcolor=blue,linktocpage=true]{hyperref}

\begin{document}

\title{Cross-Embodiment Motion Planning by Coupling Body-Agnostic World Model with Differentiable Ego-Dynamics}

\author{Zhidong Wang$^{*}$, Zirui Li$^{*}$, Haoran Xu, Yujie Yan, Zitong Shan, Jingsong Liang, Han Yu, Chen Lv%

\thanks{This work was supported in part by Nanyang Technological University. $^{*}$Zhidong Wang and Zirui Li contributed equally to this work. (\textit{Corresponding author: Chen Lv.})}
\thanks{Zhidong Wang, Zirui Li, Haoran Xu, Yujie Yan, Zitong Shan, Jingsong Liang and Chen Lv are with the School of Mechanical and Aerospace Engineering, Nanyang Technological University, Singapore
(e-mail: \texttt{zhidong001@e.ntu.edu.sg};
\texttt{zirui.li@ntu.edu.sg};
\texttt{hxu028@e.ntu.edu.sg};
\texttt{yujie018@e.ntu.edu.sg};
\texttt{zitong.shan@ntu.edu.sg};
\texttt{jingsong002@e.ntu.edu.sg};
\texttt{lyuchen@ntu.edu.sg}).

Zhidong Wang is also with the Collaborative Initiative, Interdisciplinary Graduate Programme, Nanyang Technological 
University, Singapore.

Han Yu is with the College of Computing and Data Science, Nanyang Technological University, Singapore (e-mail: \texttt{han.yu@ntu.edu.sg}).}}

\maketitle

\begin{abstract}
Motion planning in clutter is embodiment-specific: the feasible
free space and reachable commands depend on each robot's geometry and
dynamics. Existing cross-embodiment policies output
normalized commands that disregard these differences, or
compensate through per-robot training and distillation.
World models can in principle pool cross-embodiment data, yet
command-conditioned models bind their latent dynamics to a
single action space, while pose-conditioned models omit the
mapping from commands to body-specific motion.
On flat ground, however, embodiment affects the planner's observations only through the camera ego-motion.
This letter proposes \method{}, an hourglass-shaped architecture that
factorizes cross-embodiment motion planning at this interface: a
body-agnostic world model and each robot's differentiable ego-dynamics,
which map commands to that ego-motion, are coupled only through it.
The world model is trained on data pooled across embodiments and
camera mounts; each robot's ego-dynamics is a reduced-order template
with identified physical parameters.
A generalist parameter-conditioned policy is learned once at this junction:
each imagined step rolls out the template and then the world model,
and imagination training through each robot's template makes the
policy body-aware.
On seven simulated robots of five locomotion types, the generalist policy
outperforms imitation-learning and diffusion-policy baselines on the
same world model by 3.8--10.6
percentage points in success rate and matches per-robot specialists
without dedicated training, while approaching its parameter-conditioned
MPPI teacher within 1.4--2.8 points without an online map.
After three minutes of teleoperated driving for identification,
the synthetic-data-trained model transfers zero-shot to a real
skid-steer robot unseen in training, navigating unseen indoor
scenes.
\end{abstract}


\section{Introduction}
\label{sec:intro}
\IEEEPARstart{M}{obile} robots vary widely in embodiment,
from wheeled bases and quadrupeds to humanoids and
wheeled-legged platforms, yet they share the core problem of
local motion planning: generating velocity commands that steer
the robot toward a nearby goal through cluttered environments
without collision. Training a separate learned planner for each
embodiment cannot scale, motivating recent cross-embodiment approaches
based on foundation
models~\cite{10610665,pmlr-v270-doshi25a,ICLR2026_ceb45c48},
embodiment-aware
policies~\cite{11244111,11696012,ICLR2026_514bdfb1}, and world
models~\cite{11128692,11093950}. Global planning produces a route that is largely
embodiment-agnostic, whereas local planning must respect each
robot's geometry and dynamics: the geometry constrains which
gaps are passable, and the dynamics govern which commands can
be executed in time. A cross-embodiment local planner must
therefore account for these body-specific properties without
requiring a dedicated training stage or dataset for each robot.

\begin{figure}[!t]
\centering
\includegraphics[width=\columnwidth]{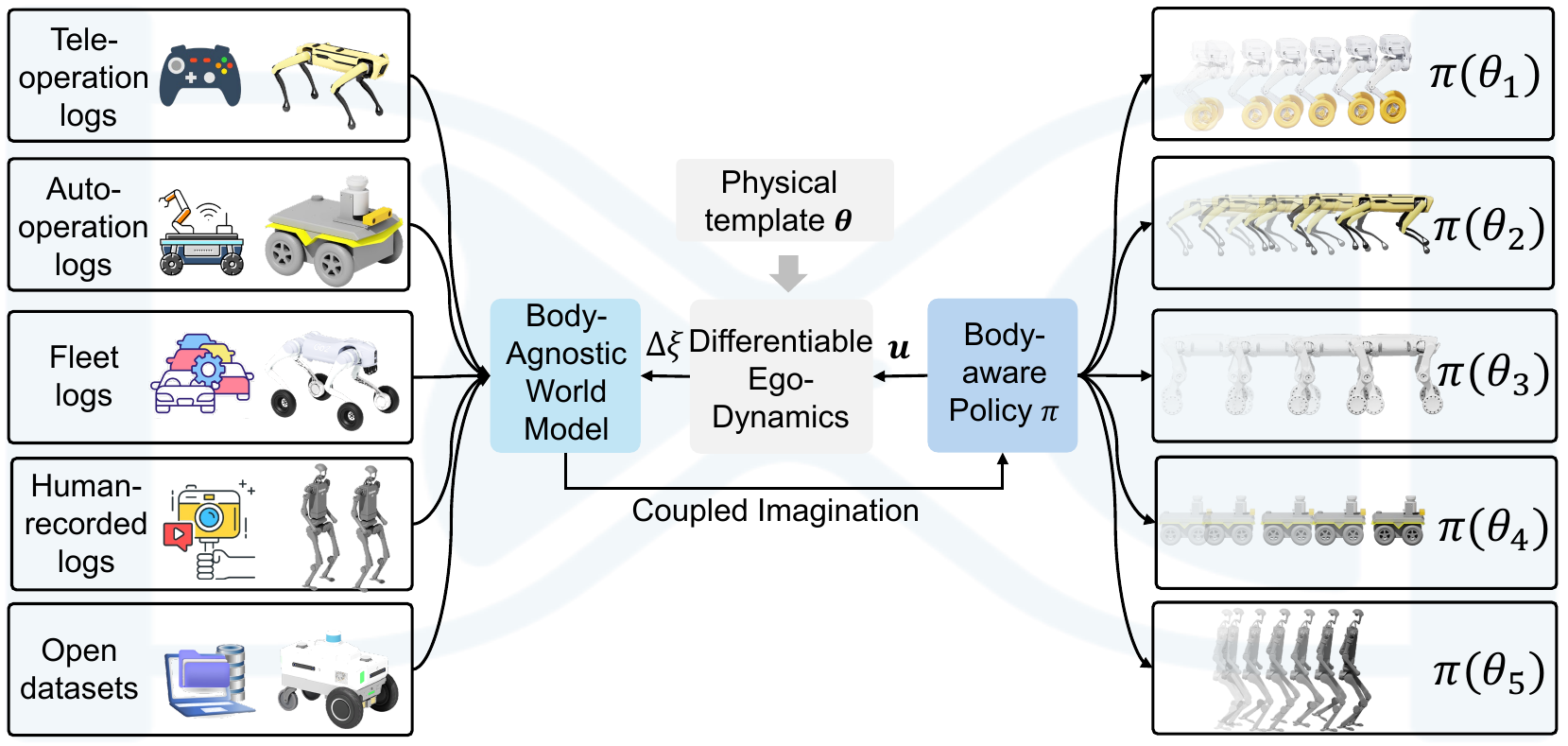}
\caption{\method{}: a body-agnostic world model and the differentiable ego-dynamics with physical parameters $\vth$, which map the commands $\vu$ to the camera ego-motion $\dxi$, are joined only through $\dxi$; a cross-embodiment generalist policy $\pi$ is learned at this junction.}
\label{fig:teaser}
\end{figure}

\begin{figure*}[!t]
\centering
\includegraphics[width=1.0\textwidth]{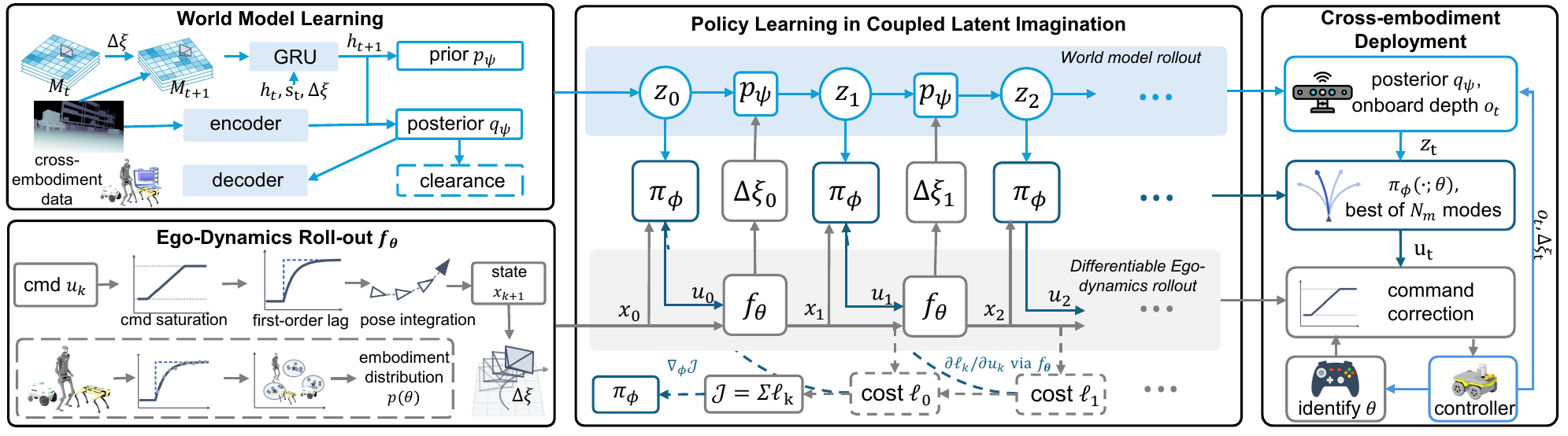}
\caption{Overview of \method{}. Left: the body-agnostic world model $p_\psi$, and the differentiable ego-dynamics $f_{\vth}$ with physical parameters $\vth$. Middle: in coupled latent imagination, the two modules interact through the ego-motion $\dxi_k$, and a single $\vth$-conditioned policy is trained by the analytic gradient of $J=\sum\ell_k$. Right: deployment with the identified $\vth$.}
\label{fig:framework}
\end{figure*}

Existing approaches handle these embodiment differences at
different levels, but each conflates the environment
representation with the embodiment in some form.
Shared-interface policies output normalized commands and
delegate execution to each robot's low-level
controller~\cite{10610665,11696748,ICLR2026_ceb45c48}, leaving
the planner blind to the robot's dynamics.
Embodiment-aware policies restore dynamics awareness through
per-robot training, whether by distilling
specialists~\cite{11244111,11696012} or by attaching residual
modules trained with reinforcement
learning~\cite{11696012,ICLR2026_514bdfb1}, so each new robot
needs its own training stage.
World models move the problem into the
representation: command-conditioned models need
command-labelled data and retraining for each
robot~\cite{11128692,11696012}, while pose-conditioned models
pool data but drop the mapping from commands to
motion~\cite{11093950,10.1007/978-3-032-37372-4_21}.
All three lines of work entangle the environment with the
embodiment, yet on flat ground the embodiment affects observations
only through the camera ego-motion.

Building on this observation, this letter proposes \textbf{\method{}},
an hourglass-shaped architecture (Fig.~\ref{fig:teaser}) that
factorizes cross-embodiment motion planning at the camera ego-motion: a
body-agnostic world model and the differentiable ego-dynamics of each
robot, which map commands to this ego-motion, are coupled only through
it. A parameter-conditioned policy is learned at this junction
(Fig.~\ref{fig:framework}). The main contributions are:
\begin{itemize}
    \item \textbf{Body-agnostic world model.} Conditioned only on
    ego-motion and camera parameters, the world model is body-agnostic,
    pooling data across embodiments and camera mounts, and is nearly
    unaffected when a robot's own data are held out, provided its camera
    mount is covered.
    \item \textbf{Cross-embodiment policy through differentiable
    ego-dynamics.} At each imagined step, coupled imagination rolls out each robot's
    template and then the world model, training a generalist policy
    by analytic gradients and making it body-aware without per-robot
    specialists or dedicated training.
    \item \textbf{Cross-embodiment evaluation and zero-shot transfer.}
    On seven simulated robots of five locomotion types, the generalist
    policy outperforms imitation-learning and diffusion-policy baselines
    on average and matches per-robot specialists; after three minutes of teleoperated
    driving for identification, the synthetic-data-trained model transfers
    zero-shot to an unseen real skid-steer robot.
\end{itemize}

\section{Related Work}
\label{sec:related}

Cross-embodiment policies for navigation and local planning take two
broad forms.
Shared-interface methods map all robots onto a common output space such
as normalized waypoints or mid-level motion
commands~\cite{10610665,pmlr-v270-doshi25a,cheng2025navila,11278012,11696748,ICLR2026_ceb45c48,ICLR2026_125be378,11610867}
and delegate execution to each robot's low-level controller. The planner is therefore unaware of each robot's dynamics.
Embodiment-aware methods incorporate dynamics through training:
randomizing embodiment
parameters~\cite{11696447}, distilling per-robot or per-embodiment-family
specialists~\cite{11244111}, attaching per-robot residual or refinement
modules~\cite{11696012,ICLR2026_514bdfb1,11692608}, or conditioning on
footprint and velocity limits~\cite{10900448}. Most methods require
a dedicated training stage or simulation
model for every new robot, and none of them incorporates a
differentiable model of each robot's dynamics into policy training.

World models for cross-embodiment navigation condition either on
commands or on relative pose. Command-conditioned models learn
environment and dynamics jointly, so they need command-labelled data
and retraining for every robot~\cite{11128692}; adding per-robot residual policies on top
facilitates deployment but inherits this
limitation~\cite{11696012}. Pose-conditioned models pool data across
robots by conditioning on
ego-motion~\cite{11093950,10.1007/978-3-032-37553-7_16,10.1007/978-3-032-37372-4_21},
but omit the mapping from commands to motion,
leaving the policy unaware of body-specific dynamics. How to keep the world model body-agnostic while exposing each robot's dynamics to the
policy remains open.

\section{Methodology}
\label{sec:method}

Fig.~\ref{fig:framework} outlines \method{}. The system addresses
local planning: given a goal $\vg_t$ in the body frame and an optional
route from a global planner, each robot observes depth images
$\vo_t$ and ego-motion estimates, and outputs body-frame velocity
commands $\vu_t=(u_v,u_\omega)$ to its low-level controller at a fixed
decision interval $\Delta t$. Each robot is parameterized by
physical parameters $\vth$ (speed and acceleration limits, response
time constants, gait frequency and footprint) and camera parameters
$\vc$ (mounting pose and field of view), denoted $\vc_t$ at time~$t$.
The mount defines $T_c\in SE(2)$, placing the camera ground pose
$\xi^c_t=\xi_tT_c$ relative to the body pose $\xi_t$; the camera
ego-motion is $\dxi_t=(\xi^c_t)^{-1}\xi^c_{t+1}$. The state $\vx_t$
comprises the pose, velocity, and low-level controller state.

On flat ground, with out-of-plane camera motion (e.g., gait-induced
pitch, roll, and height oscillation) treated as noise, the observation
depends on the environment geometry $\mathcal{S}_t$ and the camera
ground pose, $\vo_t=\Gamma(\mathcal{S}_t,\xi^c_t;\vc_t)$ up to noise. This
dependence is fully characterized by the camera ego-motion $\dxi_t$ and
the camera parameters $\vc$; the ego-motion itself is determined by the
robot's dynamics, which map the current state and command to the next
pose, and by the mount $T_c$. The observation model therefore factorizes at the ego-motion into
an environment factor $p_W$ and an ego-dynamics factor $p_E$:
\begin{equation}
\begin{aligned}
&p\big(\vo_{t+1}\mid\vo_{\le t},\dxi_{<t},\vx_t,\vu_t;\vth,\vc\big)\\
&\quad=\!\int\! p_W\big(\vo_{t+1}\mid\vo_{\le t},\dxi_{\le t};\vc\big)\,p_E\big(\dxi_t\mid\vx_t,\vu_t;\vth,\vc\big)\,\mathrm{d}\dxi_t.
\end{aligned}
\label{eq:factor}
\end{equation}

\begin{remark}
Given $\dxi_{\le t}$ and $\vc$, the next observation is independent of
commands and $\vth$. A world model conditioned only on $\dxi$ and
$\vc$ is thus body-agnostic by construction. A
command-conditioned world model instead marginalizes
$p_E(\cdot;\vth,\vc)$ into Eq.~\eqref{eq:factor}, binding it to seen
embodiments. A new embodiment changes $(\vth,\vc)$ rather than $p_W$,
requiring only identification of $\vth$ and no
new world-model data, provided its camera mount is covered.
\end{remark}

The objective is to learn a $\vth$-conditioned policy
$\pi_\phi(\cdot;\vth)$ over the embodiment distribution $p(\vth)$,
\begin{equation}
\min_{\phi}\mathbb{E}_{p(\vth)}\big[J(\phi;\vth)\big],\ \,
J(\phi;\vth)=\mathbb{E}\big[\textstyle\sum_{k=0}^{H-1}
\ell(\vx_{k+1},\vu_k;\vth,\Geo)\big],
\label{eq:objective}
\end{equation}
where $\ell$ is the planning cost of step $k$ and trajectories of $H$ steps
(horizon $T_h=H\Delta t$) are rolled out by $\pi_\phi(\cdot;\vth)$. The cost and
the policy depend on the embodiment only through $(\vth,\vc)$, so
deploying on a new robot requires only $\hat{\vth}$, the command-response
correction $\hat{\delta}$ (Sec.~\ref{subsubsec:identification}), and its camera
parameters. During training,
the cost $\ell$ may use the scene geometry $\Geo$ (a Euclidean
signed distance field and the geodesic cost-to-go), reconstructed
offline from logs or taken from a metric map. At deployment, the policy
uses only onboard depth, ego-motion, and
the goal from a global planner.

The factorization in Eq.~\eqref{eq:factor} structures the architecture
around three requirements: \emph{body-agnosticism} of the world model,
\emph{explicit dynamics} of each robot, and their \emph{coupling} in
latent imagination~\cite{DBLP:conf/iclr/HafnerLB020}, where the policy is trained. The
world model and the ego-dynamics are coupled only through the
ego-motion $\dxi_t\in SE(2)$, given the camera parameters $\vc$.

\subsection{Body-Agnostic World Model}
\label{subsec:wm}

To pool data across robots, the world model is conditioned only on camera
ego-motion $\dxi_t$ and camera parameters $\vc$, independent of which
robot produced that motion. The learned model $p_\psi\approx p_W$ is a
recurrent state-space model~\cite{pmlr-v97-hafner19a} with state
$\vz_t=(\vh_t,\vs_t,\vM_t)$: a deterministic recurrent state, a
stochastic latent variable, and an egocentric spatial memory.

\subsubsection{Ego-Motion Conditioning}
\label{subsubsec:egomotion}
The state transition and the rigid transformation of the memory are conditioned only on the
camera ego-motion:
\begin{equation}
\begin{aligned}
\vz_{t+1}&\sim p_\psi\big(\vz_{t+1}\mid\vz_t,\dxi_t\big),\\
\vM^-_{t+1}(\vp)&=\vM_t\big(\dxi_t\,\vp\big)
  \quad\forall\,\vp\in\mathbb{R}^2,
\end{aligned}
\label{eq:wm}
\end{equation}
where $\vp$ is a coordinate in the camera ground frame and
$\vM^-_{t+1}$ is the memory before the next observation is incorporated.
The posterior
$q_\psi(\vz_{t+1}\mid\vz_t,\dxi_t,\vo_{t+1};\vc_{t+1})$ fuses the
next observation, encoded jointly with the camera parameters. Commands,
robot identities, and $\vth$ are excluded from the model. As a metric
increment of the camera ground pose, $\dxi$ represents the same
displacement regardless of the robot, and the rigid transformation of $\vM$
is parameter-free.

\subsubsection{Egocentric Spatial Memory}
\label{subsubsec:memory}
The spatial memory $\vM_t$ is a multi-channel grid of latent features centered at the
camera, rigidly transformed by each ego-motion
step~\cite{10.1007/978-3-030-67661-2_31} (Fig.~\ref{fig:framework}, top left). During posterior inference,
depth points projected with the camera intrinsics and mount $\vc_t$ are
fused into the memory through a convolutional gate; during imagination, only the rigid transformation is applied. The recurrent
update receives a readout of $\vM_t$, the depth features are combined
with an embedding of $\vc_t$, and the depth decoder is conditioned
on $\vc_t$.

\subsubsection{Training Data From Heterogeneous Robots}
\label{subsubsec:anyrobot}
The world model is trained on sequences
$\mathcal{D}=\{(\vo_t,\dxi_t,\vc_t)\}$ of depth images, camera
ego-motion (from any pose source), and camera parameters. Logs from robots with different dynamics and camera mounts,
whether teleoperated, autonomous, or handheld, are valid samples of
$p_W$: the relevant diversity lies in ego-motions and viewpoints,
not in the number of robots.

The training loss consists of depth reconstruction, an optional semantic
target, the KL divergence, open-loop prior roll-outs
with latent overshooting~\cite{pmlr-v97-hafner19a}, and clearance prediction from the latent state
and the spatial memory. Clearance is supervised by $\Geo$ on cells observed within
the training window, and no loss term depends on the embodiment dynamics.

\subsection{Differentiable Ego-Dynamics}
\label{subsec:dyn}

For the generalist policy, each robot's dynamics must be
differentiable and appear in the policy gradient, without requiring a
learned dynamics model or per-robot training. A reduced-order template
provides this throughout the pipeline: it rolls out each robot's
dynamics in imagination, generates trajectory candidates for the
planner that initializes the policy, and projects commands onto the admissible command set at deployment.

\subsubsection{Templates and Anchors}
\label{subsubsec:templates}
Each robot executes velocity commands through its own low-level
controller (a vendor controller or a learned locomotion policy). On the
planning time scale $\Delta t$, this closed-loop system, termed the
\emph{anchor}, is approximated by a reduced-order
template~\cite{10.1242/jeb.202.23.3325},
\begin{equation}
\begin{aligned}
\vx_{k+1}&=f_{\vth}\big(\vx_k,\Pi_{\Uth}(\vu_k)\big),\\
\big\|\xi^a_k-\xi_k\big\|&\le\varepsilon_{\vth}(k\Delta t),
  \quad k\Delta t\le T_h,
\end{aligned}
\label{eq:template}
\end{equation}
where $\Pi_{\Uth}$ projects the command onto the admissible command set $\Uth$,
$\xi^a_k$ is the anchor pose under the same command sequence, and
$\varepsilon_{\vth}$ is the approximate-simulation
bound~\cite{9310228}. A single template structure is used for all five
locomotion types; its template error $\varepsilon_{\vth}$ varies across robots (Table~\ref{tab:robots}).

\subsubsection{The Template \texorpdfstring{$f_{\vth}$}{f}}
\label{subsubsec:template}
The projection $\Pi_{\Uth}$ enforces kinematic and dynamic constraints:
speed limits, a lateral-acceleration bound, and, for legged robots, a
slew-rate limit on each velocity axis proportional to half the
gait frequency, approximating the closed-loop bandwidth of the
low-level locomotion controller. The template propagates the state
through first-order lag dynamics with time constants $\tau_v$ and
$\tau_\omega$ and acceleration limits, and integrates the nonholonomic
kinematics over $N_s$ sub-steps (Fig.~\ref{fig:framework}, bottom left). The footprint $\mathcal{F}_{\vth}$
consists of circles determined by the robot's length and width.

\subsubsection{Identification and the Embodiment Distribution}
\label{subsubsec:identification}
Speed limits and robot dimensions in $\hat{\vth}$ are obtained from the
datasheet or direct measurement. The response time constants and
acceleration limits are fitted to velocity step responses recorded
during a few minutes of identification maneuvers. The gait
frequency is extracted from the vertical oscillation of the robot's
body. The entire identification is data-driven and requires no simulation
or per-robot training.
With $\hat{\vth}$ fixed, a command-response gain and offset
$\hat{\delta}$, an affine map from the commanded to the executed
velocity, are fitted to the executed motion and inverted as a
feedforward correction at deployment. The template error
$\varepsilon_{\vth}$ is quantified as a high-percentile bound on the
maximum footprint deviation within the planning horizon $T_h$.
The embodiment distribution $p(\vth)$ is constructed by augmenting each
training robot's identified parameters with log-uniform perturbations
of size and dynamics, forming a parameter neighborhood around each robot.

\subsection{Cross-Embodiment Policy Learning in Coupled Imagination}
\label{subsec:policy}

\begin{algorithm2e}[t]
\small
\DontPrintSemicolon
\SetAlgoLined
\KwIn{frozen world model $p_\psi$; posterior states $\mathcal{D}_{\vz}$ inferred on $\mathcal{D}$; embodiment distribution $p(\vth)$; template $f_{\vth}$; scene geometry $\Geo$; horizon $H$; modes $N_m$}
\KwOut{generalist policy $\pi_\phi$}
Initialize $\pi_\phi$ from a $\vth$-conditioned sampling-based planner on $\mathcal{D}_{\vz}$; store as $\pi_{\phi_0}$.\;
\While{not converged}{
  Sample $(\vz_0,\vg)\sim\mathcal{D}_{\vz}$, $\vth\sim p(\vth)$ with matching camera mount; set $\vx_0$.\;
  \For(\tcp*[f]{in parallel}){$m=1$ \KwTo $N_m$}{
    \For{$k=0$ \KwTo $H-1$}{
      $\vu_k\leftarrow\pi_\phi^{(m)}(\vz_k,\vx_k;\vth)$\;
      $\vx_{k+1}\leftarrow f_{\vth}\big(\vx_k,\Pi_{\Uth}(\vu_k)\big)$\tcp*{dynamics}
      $\dxi_k\leftarrow(\xi^c_k)^{-1}\xi^c_{k+1}$\;
      $\vz_{k+1}\leftarrow\mu_\psi(\vz_k,\dxi_k)$\tcp*{world model}
    }
    $J^{(m)}\leftarrow\sum_{k=0}^{H-1}\ell(\vx_{k+1},\vu_k;\vth,\Geo)$.\;
  }
  $m^\star\!\leftarrow\!\arg\min_m J^{(m)}$; relaxed WTA: weight $m^\star$ by $1\!-\!\eta$, rest by $\tfrac{\eta}{N_m-1}$.\;
  Add critic loss on mode costs and trust-region regularizer toward $\pi_{\phi_0}$.\;
  Update $\phi$ via analytic gradient~\eqref{eq:grad} through $f_{\vth}$ and $\mu_\psi$.\;
}
\caption{Cross-embodiment policy learning.}
\label{alg:coupled}
\end{algorithm2e}

\subsubsection{Coupled Latent Imagination}
\label{subsubsec:coupled}
Each imagination episode starts from a world-model state $\vz_0$
sampled from logged sequences and paired with an embodiment
$\vth\sim p(\vth)$ whose camera mount is consistent with $\vz_0$. At
each step $k$, the policy outputs a command, the template rolls out
the robot's dynamics to produce $\dxi_k$, and the world-model prior
predicts the next latent state (Fig.~\ref{fig:framework}, middle):
\begin{equation}
\begin{aligned}
\vu_k&=\pi_\phi\big(\vz_k,\vx_k;\vth\big),
  &\vx_{k+1}&=f_{\vth}\big(\vx_k,\Pi_{\Uth}(\vu_k)\big),\\
\dxi_k&=\big(\xi^c_k\big)^{-1}\xi^c_{k+1},
  &\vz_{k+1}&=\mu_\psi\big(\vz_k,\dxi_k\big),
\end{aligned}
\label{eq:imagine}
\end{equation}
where $\mu_\psi$ is the prior mean (including the rigid transformation of
the spatial memory), and the goal, route, and velocity inputs to the policy are
derived from $\vx_k$. Eq.~\eqref{eq:imagine} instantiates the
factorization in Eq.~\eqref{eq:factor} with the template as $p_E$ and
the prior as $p_W$: the robot's dynamics determine the ego-motion, and
the world model predicts the next observation. Within a batch, episodes
differ only in $\vth$ and their initial states.

\subsubsection{Objective and Gradient Paths}
\label{subsubsec:gradient}
The cost $\ell$ penalizes collision between the robot's footprint
$\mathcal{F}_{\vth}$ and obstacles in the scene geometry $\Geo$
(represented as a Euclidean signed distance field), evaluated at each
sub-step pose, and violations of a speed-dependent clearance margin
within the stopping distance. It rewards progress along the geodesic
cost-to-go, penalizes low velocity in obstacle-free regions, and
includes terms for heading alignment, goal reaching at the horizon end,
route adherence, and command smoothness. With a deterministic template
and the prior mean, the analytic policy gradient follows the recursion
{\allowdisplaybreaks
\begin{align}
\frac{\mathrm{d}J}{\mathrm{d}\phi}&=\sum_{k=0}^{H-1}
  \Big(\frac{\partial\ell_k}{\partial\vx_{k+1}}
  \frac{\mathrm{d}\vx_{k+1}}{\mathrm{d}\phi}
  +\frac{\partial\ell_k}{\partial\vu_k}
  \frac{\mathrm{d}\vu_k}{\mathrm{d}\phi}\Big), \notag\\
\frac{\mathrm{d}\vx_{k+1}}{\mathrm{d}\phi}&=
  \frac{\partial f_{\vth}}{\partial\vx_k}
  \frac{\mathrm{d}\vx_k}{\mathrm{d}\phi}
  +\frac{\partial f_{\vth}}{\partial\vu_k}
  \frac{\mathrm{d}\vu_k}{\mathrm{d}\phi}, \notag\\
\frac{\mathrm{d}\vu_k}{\mathrm{d}\phi}&=
  \frac{\partial\pi_\phi}{\partial\phi}
  +\frac{\partial\pi_\phi}{\partial\vx_k}
  \frac{\mathrm{d}\vx_k}{\mathrm{d}\phi}
  +\frac{\partial\pi_\phi}{\partial\vz_k}
  \frac{\mathrm{d}\vz_k}{\mathrm{d}\phi}, \notag\\
\frac{\mathrm{d}\vz_{k+1}}{\mathrm{d}\phi}&=
  \frac{\partial\mu_\psi}{\partial\vz_k}
  \frac{\mathrm{d}\vz_k}{\mathrm{d}\phi}
  +\frac{\partial\mu_\psi}{\partial\dxi_k}
  \frac{\mathrm{d}\dxi_k}{\mathrm{d}\phi}, \label{eq:grad}
\end{align}}
with $\ell_k=\ell(\vx_{k+1},\vu_k;\vth,\Geo)$, zero initial
sensitivities, and $\mathrm{d}\dxi_k/\mathrm{d}\phi$ given by those of
$\vx_k$ and $\vx_{k+1}$ through $T_c$. The gradient propagates
along two paths: the \emph{dynamics path}, through the template
Jacobians $\partial f_{\vth}/\partial\vx_k$ and
$\partial f_{\vth}/\partial\vu_k$ and the policy's dependence on
$\vx_k$; and the \emph{world-model path}, through the prior, as
earlier commands alter $\dxi_{<k}$ and thereby the latent state $\vz_k$
that the policy reads at later steps. Since $\ell$ does not depend on
$\vz$, the world model enters $J$ only through these policy inputs.
The two paths join at $\dxi$.

\begin{remark}
The parameters $\vth$ enter the policy gradient through the template
Jacobians and the footprint in $\ell$. The Jacobians encode the
response time constants, while the acceleration and slew-rate limits zero
the gradient of commands that saturate them.
Under perfect command tracking, these Jacobians reduce to one
kinematic integrator for all robots, and $\vth$ would enter only through the
policy input and the footprint. With the reduced-order template, the same environment and
goal produce $\vth$-dependent gradients, making the policy
body-aware. The world-model path is shared by all robots.
\end{remark}

\subsubsection{Policy Architecture and Training}
\label{subsubsec:training}
The policy reads the recurrent state $\vh_k$ and the
stochastic latent variable $\vs_k$ of $\vz_k$ (the spatial memory enters
through the recurrent readout), the route, goal, and velocity, as well as
normalized features of the robot's physical and camera parameters. To
handle multimodal planning decisions (e.g., passing an obstacle on
either side), the policy outputs $N_m$ command modes together with a
critic head. All modes are rolled out in imagination, and the mode with
the lowest cost receives weight $1-\eta$ while the remaining modes
share $\eta$ (relaxed winner-takes-all, WTA). The critic regresses each
mode's imagined cost (Algorithm~\ref{alg:coupled}).

The policy is initialized by imitation learning from a
$\vth$-conditioned sampling-based model predictive path integral (MPPI)
planner~\cite{MACENSKI2023104493} that rolls out candidates through the
same template using the portion of $\Geo$ observed along the logged
camera path. During imagination training, a trust-region regularizer
penalizes the winning mode's command deviation from this initialization.
The initialization planner is itself
$\vth$-conditioned, not a per-robot specialist; the regularizer does
not bound the deviation, and imagination training optimizes the full
policy end-to-end, including mode selection.

\subsubsection{Deployment}
\label{subsubsec:deployment}
At deployment, the world model performs posterior inference on onboard
depth without access to $\Geo$ or a global map. The ego-motion $\dxi$
is estimated by fusing odometry with the template's predicted motion
for the last command. At each decision step $\Delta t$, the command of
the mode with the lowest critic-predicted cost is corrected by the command-response gain and offset
$\hat{\delta}$, projected onto the admissible command set $\Uth$, and
sent to the low-level controller.

\section{Experiments}
\label{sec:exp}

The experiments evaluate the generalist policy across embodiments, the factorization, each component, unseen robots under template error, and the transfer to a real robot.

\subsection{Experimental Setup and Baselines}
\label{subsec:setup}

\begin{table}[!t]
\caption{Robots, Template Parameters and Method Parameters}
\label{tab:robots}
\centering
\scriptsize
\setlength{\tabcolsep}{2.4pt}
\begin{tabular}{@{}llcccccc@{}}
\toprule
Robot & Type & $v_{\max}$ & $\omega_{\max}$ & $\tau_v$ / $\tau_\omega$ & $L\times W$ & Cam. & $\varepsilon_{\vth}$\\
 & & (m/s) & (rad/s) & (s) & (m) & (m) & (m)\\
\midrule
\multicolumn{8}{@{}l}{\emph{Training}}\\
Carter & wheeled & 1.50 & 2.00 & 0.01 / 0.01 & $0.77\times0.62$ & 0.10 & 0.04\\
Spot & quadrupedal & 1.50 & 2.00 & 0.24 / 0.09 & $0.87\times0.44$ & 0.10 & 0.33\\
H1 & humanoid & 1.00 & 1.00 & 0.35 / 0.07 & $0.44\times0.56$ & 0.15 & 0.19\\
G1 & humanoid & 1.00 & 1.00 & 0.48 / 0.05 & $0.48\times0.40$ & 0.10 & 0.27\\
M20 & wheeled-legged & 1.50 & 1.57 & 0.01 / 0.10 & $0.89\times0.69$ & 0.39 & 1.62\\
Go2-W & wheeled-legged & 1.00 & 1.00 & 0.13 / 0.03 & $0.73\times0.47$ & 0.35 & 0.45\\
TRON1 & biped. wheel-leg. & 0.70 & 1.00 & 0.01 / 0.18 & $0.42\times0.53$ & 0.18 & 1.14\\
\midrule
\multicolumn{8}{@{}l}{\emph{Unseen in training (zero-shot)}}\\
Jackal & skid-steer & 1.50 & 2.50 & 0.01 / 0.20 & $0.48\times0.42$ & 0.25 & 3.98\\
iw.hub & wheeled & 1.50 & 2.50 & 0.01 / 0.14 & $1.42\times0.65$ & 0.37 & 1.70\\
Scout Mini & skid-steer (real) & 1.00 & 0.52 & 0.20 / 0.17 & $0.62\times0.65$ & 0.21 & 0.08\\
\midrule
\multicolumn{8}{@{}l}{\emph{Parameters of \method{}}}\\
\multicolumn{2}{@{}l}{Decision interval $\Delta t$} & \multicolumn{2}{l}{0.2\,s} & \multicolumn{2}{l}{Horizon $T_h$} & \multicolumn{2}{l}{5\,s}\\
\multicolumn{2}{@{}l}{Template sub-steps $N_s$} & \multicolumn{2}{l}{4} & \multicolumn{2}{l}{Footprint circles} & \multicolumn{2}{l}{3}\\
\multicolumn{2}{@{}l}{Depth image} & \multicolumn{2}{l}{$144\times256$} & \multicolumn{2}{l}{Camera parameters $\vc$} & \multicolumn{2}{l}{$\mathbb{R}^7$}\\
\multicolumn{2}{@{}l}{Spatial memory $\vM$} & \multicolumn{2}{l}{$48\times48\times8$} & \multicolumn{2}{l}{Grid resolution} & \multicolumn{2}{l}{0.15\,m}\\
\multicolumn{2}{@{}l}{Depth encoder} & \multicolumn{2}{l}{ResNet-18} & \multicolumn{2}{l}{State dims.\ $\vh$ / $\vs$} & \multicolumn{2}{l}{1024 / 512}\\
\multicolumn{2}{@{}l}{Policy modes $N_m$} & \multicolumn{2}{l}{4} & \multicolumn{2}{l}{Relaxed WTA $\eta$} & \multicolumn{2}{l}{0.05}\\
\multicolumn{2}{@{}l}{Embodiment features} & \multicolumn{2}{l}{12} & \multicolumn{4}{l}{$p(\vth)$: size $\pm4$\%, dyn.\ $\pm10$--$25$\%}\\
\bottomrule
\end{tabular}
\begin{flushleft}
\scriptsize Cam.: forward offset of the camera from the rotation center. $\varepsilon_{\vth}$: 95th-percentile bound on the maximum footprint deviation of the template over 5\,s; $\tau_v$ and $\tau_\omega$ include the measured pure delay; $\varepsilon_{\vth}$ is the 95th-percentile 5\,s position error from real-robot data under the clipped commands.
\end{flushleft}
\end{table}

\begin{figure}[!t]
\centering
\includegraphics[width=0.9\columnwidth]{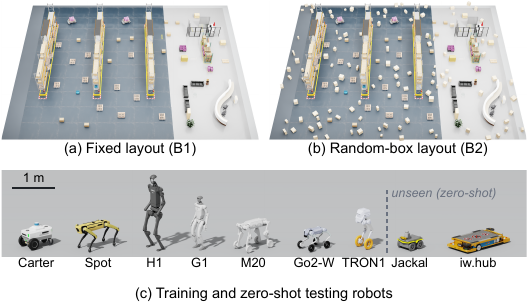}
\caption{Benchmarks and robots: (a) fixed layout B1 and (b) random-box layout B2 of the mixed indoor scene; (c) training and zero-shot-test robots.}
\label{fig:env}
\end{figure}

\subsubsection{Robots and Benchmarks}
\label{subsubsec:robots}
Seven training robots of five locomotion types are simulated in Isaac
Sim with their own low-level controllers, two additional wheeled robots
are unseen in training, and a real robot is used in
Sec.~\ref{subsec:real} (Table~\ref{tab:robots}, Fig.~\ref{fig:env}).
All robots use a pinhole depth camera (RealSense D435). Each
episode is a local planning task toward a point goal, with a straight
reference route in place of a global planner; collision ends the
episode as a failure. An episode not reaching the goal within
25\,s ends as a timeout, without early termination for lack
of progress. Both benchmarks use a $39.7\times30.6$\,m indoor scene
mixing an office, a laboratory, and warehouse racks~\cite{11696012}. Benchmark~B1 uses its fixed layout and scores all
640 episodes per robot. Benchmark~B2 adds 8--16
random boxes per 100\,m$^2$ across 80 new layouts absent from all
training data, with eight episodes each;
this dense clutter lowers the SR of every method (Table~\ref{tab:main}).
B2 scores only episodes whose goal is reachable given the
footprint and the time limit, and is held out from all
design and tuning choices.

\subsubsection{Training Data and Parameters}
\label{subsubsec:data}
Table~\ref{tab:robots} lists the main parameters. No robot is simulated
in the world-model data: synthetic depth sequences are rendered by a
robot-free rig of virtual cameras moving through the same scene with
randomized box layouts. The rig emulates the ego-motion of diverse
dynamics: its trajectories are rolled out by the template over a
continuous parameter range that covers and extends beyond the training
robots, including each training robot's parameter neighborhood with its
$\hat{\delta}$ and tracking noise, with recorded body
oscillations superimposed; the camera mounts span heights of
0.13--1.85\,m, including cross-robot pairings of motion and mount.
$\dxi$ is computed from the rig poses, $\Geo$ from the scene geometry,
and the semantic targets from the renderer. Holding out a robot removes
all of its motions and mounts; those of the unseen robots are never
included.

\subsubsection{Baselines}
\label{subsubsec:baselines}
BC-$\theta$ is the imitation initialization of \method{} without
imagination training: a single-mode behavior-cloning
policy~\cite{11128692} on world-model latent states, whose command head
is copied into the $N_m$ modes of \method{} with small
perturbations, so that all modes start from BC-$\theta$. DP-$\theta$ is a
diffusion policy~\cite{11696748} trained on the same latents and
labels. Both are conditioned on $\vth$. DWA-$\theta$ is the
dynamic-window planner~\cite{MACENSKI2023104493} and MPPI-$\theta$ is the
sampling-based planner of Sec.~\ref{subsubsec:training}; both roll out
each robot's identified template with $\hat{\delta}$, widen the
collision margin by the template error within the stopping
horizon (0.01--0.34\,m), and plan on an occupancy map
accumulated over the episode with an online geodesic cost-to-go. Privileged
MPPI runs MPPI-$\theta$ on the full $\Geo$ as a reference. All methods use the simulator pose and one execution layer: compensation of $\hat{\delta}$, projection onto
$\Uth$, and a goal-reaching controller within 0.5\,m.

\subsubsection{Metrics}
\label{subsubsec:metrics}
The success rate (SR) is reported per robot and as a seven-robot mean.
The collision rate (Coll.), success weighted by path
length (SPL, referenced to the shortest collision-free path for the
robot's footprint), and mean duration of successful episodes (Time)
are seven-robot means. Paired per-episode differences are tested with
McNemar's test. Three training seeds of \method{}, evaluated with an
earlier execution layer, differ by at most 0.1 points on B2.

\subsection{Cross-Embodiment Motion Planning Performance}
\label{subsec:main}

\begin{table*}[!t]
\caption{Motion Planning Performance Across Seven Embodiments}
\label{tab:main}
\centering
\scriptsize
\setlength{\tabcolsep}{2.4pt}
\begin{tabular}{@{}lccccccccccc@{\hspace{6pt}}ccccccccccc@{}}
\toprule
 & \multicolumn{11}{c}{B2: random boxes, feasible episodes} & \multicolumn{11}{c}{B1: fixed layout, all episodes}\\
\cmidrule(lr){2-12}\cmidrule(l){13-23}
 & \multicolumn{8}{c}{SR (\%)} & Coll. & SPL & Time & \multicolumn{8}{c}{SR (\%)} & Coll. & SPL & Time\\
\cmidrule(lr){2-9}\cmidrule(lr){13-20}
Method & Carter & Spot & H1 & G1 & M20 & Go2-W & TRON1 & Mean & (\%) & (\%) & (s) & Carter & Spot & H1 & G1 & M20 & Go2-W & TRON1 & Mean & (\%) & (\%) & (s)\\
\midrule
BC-$\theta$~\cite{11128692} & 65.7 & 68.2 & 77.4 & 71.8 & 70.6 & 77.3 & 67.8 & 71.2$^{*}$ & 9.2 & 63.8 & 14.6 & 82.7 & 83.6 & 84.5 & 83.4 & 85.8 & 83.4 & 76.4 & 82.8$^{*}$ & 3.1 & 75.7 & 13.4\\
DP-$\theta$~\cite{11696748} & 75.4 & 76.4 & 80.1 & 76.1 & 71.2 & 78.0 & 70.8 & 75.5$^{*}$ & 7.9 & 67.8 & 14.1 & 86.7 & 87.2 & 86.4 & 85.6 & 82.8 & 85.3 & \textbf{76.7} & 84.4$^{*}$ & 4.0 & 76.9 & 13.4\\
DWA-$\theta$~\cite{MACENSKI2023104493} & 44.5 & 62.4 & 78.2 & 58.4 & 40.5 & 70.4 & 24.0 & 54.1$^{*}$ & \textbf{2.9} & 49.1 & 14.4 & 69.4 & 79.8 & 85.2 & 73.9 & 69.4 & 85.5 & 51.9 & 73.6$^{*}$ & \textbf{0.9} & 66.9 & 13.2\\
MPPI-$\theta$~\cite{MACENSKI2023104493} & \textbf{95.6} & \textbf{86.6} & \textbf{89.7} & \textbf{90.6} & 77.8 & \textbf{92.4} & 49.6 & \textbf{83.2}$^{*}$ & 3.2 & 70.1 & 13.0 & \textbf{98.8} & \textbf{95.0} & \textbf{89.7} & \textbf{94.5} & 92.8 & \textbf{95.8} & 70.5 & \textbf{91.0}$^{*}$ & 1.6 & 78.0 & 12.4\\
RL~\cite{11696012} & -- & -- & -- & -- & -- & -- & -- & -- & -- & -- & -- & 85.3 & 77.9 & 84.4 & 90.6 & -- & -- & -- & -- & -- & -- & --\\
\textcolor{gray}{Privileged MPPI~\cite{MACENSKI2023104493}} & \textcolor{gray}{98.8} & \textcolor{gray}{96.5} & \textcolor{gray}{95.1} & \textcolor{gray}{95.8} & \textcolor{gray}{78.4} & \textcolor{gray}{93.3} & \textcolor{gray}{55.7} & \textcolor{gray}{87.7$^{*}$} & \textcolor{gray}{2.2} & \textcolor{gray}{75.5} & \textcolor{gray}{12.8} & \textcolor{gray}{99.1} & \textcolor{gray}{98.1} & \textcolor{gray}{96.7} & \textcolor{gray}{96.7} & \textcolor{gray}{92.7} & \textcolor{gray}{93.0} & \textcolor{gray}{71.1} & \textcolor{gray}{92.5$^{*}$} & \textcolor{gray}{1.2} & \textcolor{gray}{81.5} & \textcolor{gray}{12.2}\\
\method{} (ours) & 83.2 & 81.4 & 84.3 & 85.4 & \textbf{85.0} & 81.9 & \textbf{71.5} & 81.8 & 8.9 & \textbf{73.0} & \textbf{12.9} & 92.3 & 93.0 & 88.3 & 89.2 & \textbf{93.0} & 84.8 & \textbf{76.7} & 88.2 & 2.7 & \textbf{81.5} & \textbf{12.1}\\
\bottomrule
\end{tabular}
\begin{flushleft}
\scriptsize  Means are computed before rounding. $^{*}$: differs from \method{} at $p<0.05$ (paired McNemar test). RL: reported B1 results, RGB camera. Gray: reference with full scene knowledge.
\end{flushleft}
\end{table*}

\begin{figure}[!t]
\centering
\includegraphics[width=0.9\columnwidth]{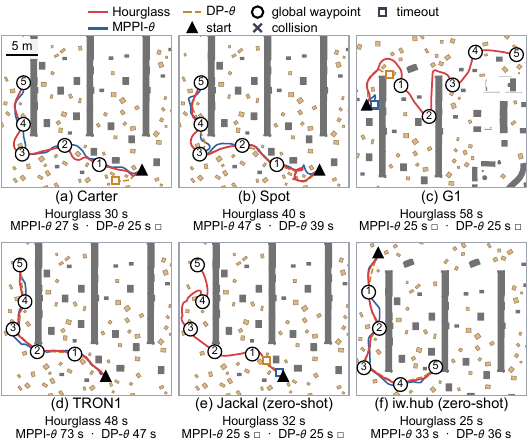}
\caption{Example trajectories through five consecutive goals. Each goal-to-goal
segment has a 25\,s time limit.}
\label{fig:tours}
\end{figure}

Table~\ref{tab:main} compares the generalist policy with the
baselines. \method{} outperforms both learned baselines
on the same world model on average and on most individual robots, and
far exceeds DWA-$\theta$. On B1, it is higher than or on par with the
reported RL results~\cite{11696012} on their four robots. MPPI-$\theta$ reaches a slightly higher mean SR, while \method{}
 achieves the highest SPL
and the shortest successful episodes among the methods without full
scene knowledge, indicating more direct but less conservative paths (Fig.~\ref{fig:tours}). MPPI-$\theta$ is also the
planner whose labels initialize \method{}
(Sec.~\ref{subsubsec:training}); as in distillation-based
cross-embodiment methods, whose generalists match their specialist
teachers~\cite{11696012}, the learned policy approaches its teacher with a
world-model update and one policy forward pass instead of an occupancy map, an online
cost-to-go, and thousands of roll-outs per step (18\,ms on a GPU vs.\ 101\,ms on a CPU for the roll-outs alone), and leads or matches it
on the robots with the largest template errors
(Sec.~\ref{subsec:newrobots}).

\subsection{Validation of the Factorization}
\label{subsec:oneworld}

\begin{figure}[!t]
\centering
\includegraphics[width=0.9\columnwidth]{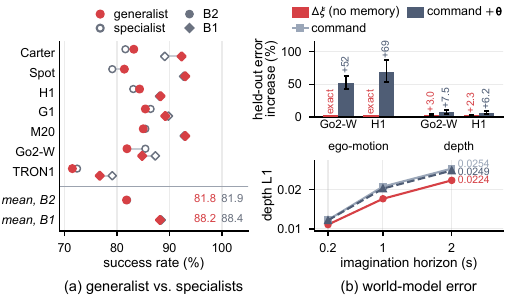}
\caption{Validation of the factorization: (a) generalist vs.\ specialists; (b) world models without spatial memory differing only in conditioning: error increase when a robot is held out (top) and imagined-depth error (bottom).}
\label{fig:oneworld}
\end{figure}

Three experiments test the factorization (Fig.~\ref{fig:oneworld}).
First, seven per-robot specialists trained with the same protocol match
the generalist policy on both benchmarks ($p=0.85$ and $0.62$), confirming that per-robot
training brings no further gain. Second, retraining the memory-based world
model without Go2-W or H1 data raises the held-out robot's imagined-depth
error by only 2.5\% and 1.6\% and leaves their pooled SR unchanged (+0.6 and
+0.2 points on B2 and B1, not significant); the posterior error
increases only for H1 (+10\%), whose camera height (about 1.5\,m) is sparsely covered once its data are removed,
indicating that generalization requires the training data to cover the
mount height, not the robot.

Third, memory-free world models compare conditioning on
$\dxi$, on commands $\vu$, and on commands plus $\vth$
(Fig.~\ref{fig:oneworld}(b)). Conditioning on $\dxi$ inherits the template's
pose accuracy and yields the lowest depth error, while conditioning on
commands and $\vth$ degrades on held-out robots. The
closed-loop SR stays within about one percentage point across the variants
(the command-conditioned model is 1.0 point higher on B1);
factoring out the ego-motion thus barely affects
closed-loop SR while producing a more accurate model that needs no
command-labelled data and generalizes across embodiments.
\subsection{Ablation Study}
\label{subsec:ablation}

\begin{table}[!t]
\caption{Ablation of the Method Components}
\label{tab:ablation}
\centering
\scriptsize
\setlength{\tabcolsep}{3pt}
\begin{tabular}{@{}lcccccc@{}}
\toprule
 & \multicolumn{3}{c}{B2} & \multicolumn{3}{c}{B1}\\
\cmidrule(lr){2-4}\cmidrule(l){5-7}
Variant & SR & Coll. & SPL & SR & Coll. & SPL\\
\midrule
\method{} & \textbf{81.8} & 8.9 & \textbf{73.0} & \textbf{88.2} & \textbf{2.7} & \textbf{81.5}\\
w/o $\vth$ (normalized commands) & 72.6$^{*}$ & 11.9 & 61.7 & 82.9$^{*}$ & 3.9 & 72.8\\
w/o egocentric spatial memory & 78.3$^{*}$ & 9.0 & 70.1 & 86.5$^{*}$ & 2.9 & 79.3\\
untrained memory clearance prediction head & 78.6$^{*}$ & 8.9 & 70.9 & 86.8$^{*}$ & 2.9 & 80.4\\
2\,s instead of 5\,s horizon & 79.9$^{*}$ & \textbf{8.5} & 72.6 & 87.5 & 2.9 & 81.2\\
\bottomrule
\end{tabular}
\begin{flushleft}
\scriptsize Seven-robot averages in \%; $^{*}$: as in Table~\ref{tab:main}. Untrained head: the memory clearance loss acts through a readout frozen at its random initialization.
\end{flushleft}
\end{table}

Table~\ref{tab:ablation} ablates each component individually; all variants
are retrained with the same protocol, and removing imagination training
reduces the policy to BC-$\theta$ (Table~\ref{tab:main}). Without $\vth$,
the policy outputs commands normalized by the robot's speed limits and
imagination uses the template with median parameters. Imagination
training and $\vth$ conditioning produce the largest gains, consistent
with the policy benefiting from embodiment-specific dynamics entering the
gradient. The spatial memory and its trained clearance prediction head add
smaller gains, and the 5\,s horizon helps mainly on B2.
\subsection{Generalization to Unseen Robots and Model Mismatch}
\label{subsec:newrobots}

\begin{figure}[!t]
\centering
\includegraphics[width=0.9\columnwidth]{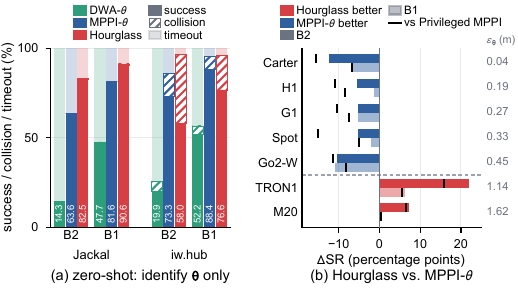}
\caption{Zero-shot transfer to unseen robots and model mismatch: (a) zero-shot outcomes; (b) SR difference to MPPI-$\theta$ and Privileged MPPI on the seven training robots, sorted by template error $\varepsilon_{\vth}$.}
\label{fig:newrobots}
\end{figure}

Deploying on an unseen robot requires only the identification of its
parameters (Fig.~\ref{fig:newrobots}(a)). On Jackal, whose size but not yaw dynamics
lie within the training range, \method{} transfers without retraining and
outperforms MPPI-$\theta$, whose failures are predominantly timeouts. On
iw.hub, whose body is longer than that of any training robot, \method{}
fails mostly by collisions and trails MPPI-$\theta$, which checks the
footprint explicitly; the transfer is thus bounded by the body sizes
covered in training.

In Fig.~\ref{fig:newrobots}(b), MPPI-$\theta$ leads on both benchmarks on the five robots with accurate
templates ($\varepsilon_{\vth}\le0.45$\,m), while on TRON1 and
M20, which have the largest $\varepsilon_{\vth}$, \method{} leads or
matches both MPPI variants. Here MPPI-$\theta$ plans with the least accurate templates
and the widest margins (0.26--0.29\,m). The imitation baselines also
exceed it on TRON1, but on M20 only \method{} does, suggesting that
imagination training contributes beyond the imitation of planner
labels; a sample of seven robots, however, is too small to establish
a correlation. On B2, the gap to MPPI-$\theta$ is driven by
collisions, while \method{} incurs fewer timeouts.

\subsection{Real-World Deployment}
\label{subsec:real}

\begin{figure}[!t]
\centering
\includegraphics[width=0.9\columnwidth]{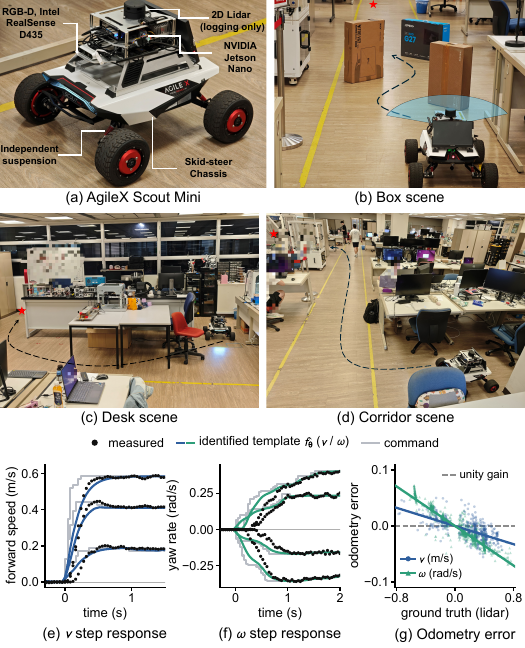}
\caption{Real robot: (a) Scout Mini, a skid-steer robot; (b--d) indoor test scenes; (e--g) dynamics identification from  teleoperated driving.}
\label{fig:realsetup}
\end{figure}

\begin{figure}[!t]
\centering
\includegraphics[width=0.9\columnwidth]{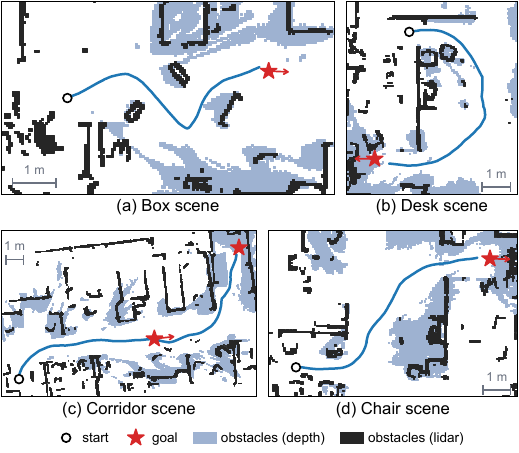}
\caption{Zero-shot deployment on the real robot.}
\label{fig:realrun}
\end{figure}

\begin{figure}[!t]
\centering
\includegraphics[width=\columnwidth]{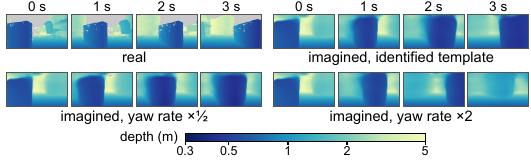}
\caption{Offline replay of the run in Fig.~\ref{fig:realrun}(a): real depth and world-model imagination roll-outs following the camera ego-motion or by counterfactual templates.}
\label{fig:realwm}
\end{figure}

This experiment demonstrates that the system, trained only on synthetic
depth data, transfers to a real robot unseen in training. The AgileX Scout
Mini (Fig.~\ref{fig:realsetup}(a)) is a skid-steer robot whose wheels
slip laterally when turning, and none of the training robots is
skid-steer; a RealSense D435 is mounted ahead of its rotation center,
and the vendor controller tracks the velocity commands. The template is
identified from three minutes of teleoperated driving with lidar scan
matching as the pose reference (Fig.~\ref{fig:realsetup}(e--g),
Table~\ref{tab:robots}): the response time constants and acceleration
limits are fitted to velocity step responses, and the command-response
gains $\hat{\delta}$ reduce to the ratios of the true to the
chassis-reported velocities, which the vendor controller tracks. On the deployment run in Fig.~\ref{fig:realrun}(a), the 95th-percentile
2\,s open-loop position error is below 4\,cm.

The world model and the policy are deployed zero-shot on an off-board
GPU with $\hat{\vth}$ from Table~\ref{tab:robots}; for safety, the
commands are further clipped to 0.25\,m/s and
0.25\,rad/s without changing $\hat{\vth}$. In the four runs shown
(Figs.~\ref{fig:realsetup}(b--d) and~\ref{fig:realrun}), covering short- and
long-range trials in four unseen indoor scenes, the robot reaches all goals without collision
or intervention. Driven only by the commands, the identified template
follows the measured speed and yaw rate over the four runs (RMS
0.016\,m/s and 0.059\,rad/s). In an offline replay of the
first run (Fig.~\ref{fig:realwm}), the world model receives no images after the start and predicts depth from the camera ego-motion of the
identified template; its 2\,s depth error is on average less than half
that of a last-frame prediction and lower at 98\% of start times.
Counterfactual templates with half or twice the yaw rate produce
visibly different imagined depth under the same commands. For each
imagined frame, the heading of its most similar real frame correlates with
the counterfactual heading ($r=0.88$), confirming that the world model
responds to the ego-dynamics.
Identification thus suffices to deploy on an unseen robot whose size
is covered in training.

\section{Conclusion}
\label{sec:conclusion}
This letter presents \method{}, which factorizes cross-embodiment
motion planning at the camera ego-motion. A generalist policy is learned once for an embodiment
distribution spanning seven robots of five locomotion types, and deploying a new robot whose size is covered in training requires
only identifying its parameters, even from simulation to
reality. The results
show that the embodiment need not be modeled in the world model, and
that per-robot training brings no further gain once it is factored out.
The main limitations are the latent clearance precision near
obstacles, the body sizes covered in training, and evaluation in a
single simulated building. Future work will address these issues with near-field geometric
representations and broader training data with real-world logs, and
will integrate global planning for long-horizon
tasks in diverse scenes with dynamic obstacles.

{\footnotesize\linespread{0.96}\selectfont
\setlength{\bibsep}{0pt plus 0.3pt}
\bibliographystyle{IEEEtranN}
\bibliography{refs}
}

\end{document}